\documentclass[runningheads]{llncs}
\usepackage{graphicx}

\usepackage{tikz}
\usepackage{comment}
\usepackage{amsmath,amssymb} 
\usepackage{color}

\usepackage{multirow}
\usepackage{bbding}

\usepackage[accsupp]{axessibility}  

\usepackage[width=122mm,left=12mm,paperwidth=146mm,height=193mm,top=12mm,paperheight=217mm]{geometry}

\begin{document}
	\pagestyle{headings}
	\mainmatter
	\def\ECCVSubNumber{1294}  
	
	\title{Complementary Bi-directional Feature Compression for Indoor 360° Semantic Segmentation with Self-distillation} 
	
	\titlerunning{CBFC}
	\author{Zishuo Zheng\inst{1,2} \and
	Chunyu Lin\inst{1,2} \and
	Lang Nie\inst{1,2}\and
	Kang Liao\inst{1,2}\and
	Zhijie Shen\inst{1,2}\and
	Yao Zhao\inst{1,2}}
	\authorrunning{Zheng et al.}
	\institute{Institute of Information Science, Beijing Jiaotong University,\\ Beijing 100044, China \and
	Beijing Key Laboratory of Advanced Information Science and Network Technology,\\ Beijing 100044, China\\
	\email{\{zszheng, cylin\}@bjtu.edu.cn}}
	\maketitle
	
	\begin{abstract}
		Recently, horizontal representation-based panoramic semantic segmentation approaches outperform projection-based solutions, because the distortions can be effectively removed by compressing the spherical data in the vertical direction.
		However, these methods ignore the distortion distribution prior and are limited to unbalanced receptive fields, e.g., the receptive fields are sufficient in the vertical direction and insufficient in the horizontal direction.
		Differently, a vertical representation compressed in another direction can offer implicit distortion prior and enlarge horizontal receptive fields.
		In this paper, we combine the two different representations and propose a novel 360° semantic segmentation solution from a complementary perspective.
		Our network comprises three modules: a feature extraction module, a bi-directional compression module, and an ensemble decoding module.
		First, we extract multi-scale features from a panorama.
		Then, a bi-directional compression module is designed to compress features into two complementary low-dimensional representations, which provide content perception and distortion prior.
		Furthermore, to facilitate the fusion of bi-directional features, we design a unique self distillation strategy in the ensemble decoding module to enhance the interaction of different features and further improve the performance.
		Experimental results show that our approach outperforms the state-of-the-art solutions with at least 10\% improvement on quantitative evaluations while displaying the best performance on visual appearance.
	\end{abstract}

	\section{Introduction}

	Panoramic images captured by omnidirectional cameras can provide a wide field-of-view (FoV), making it more practical in many crucial scene perception tasks \cite{de2018eliminating}, \cite{ma2020stage}, \cite{shen2021distortion},  \cite{zhang2021deeppanocontext}.
	As a fundamental topic in computer vision, semantic segmentation aims to assign each pixel in the image a category label and is critical for various applications such as pose estimation \cite{peng2019pvnet}, autonomous vehicles \cite{siam2018comparative}, augmented reality \cite{azuma1997survey}.
	Directly applying normal FoV semantic segmentation methods \cite{huang2019ccnet}, \cite{liu2021swin}, \cite{long2015fully}, \cite{zhao2017pyramid} to 360° images is not satisfactory due to the significant distortions in panoramas (usually produced from equirectangular projection---ERP) and large mismatch of FoV between panoramas and normal Fov images.

	To overcome the above limitations, some researchers propose to adopting different projection formats (e.g., cubemap projection and icosahedron groups) \cite{eder2020tangent}, \cite{zhang2019orientation} or spherical convolutions \cite{cohen2019gauge}, \cite{esteves2020spin}, \cite{jiang2018spherical} to decrease the negative effects of panoramic distortions.
	However, these methods sacrifice either accuracy or efficiency and fail to perceive precise panoramic structures.

	\begin{figure}[t]
		\centering
		\includegraphics[width=0.9\linewidth]{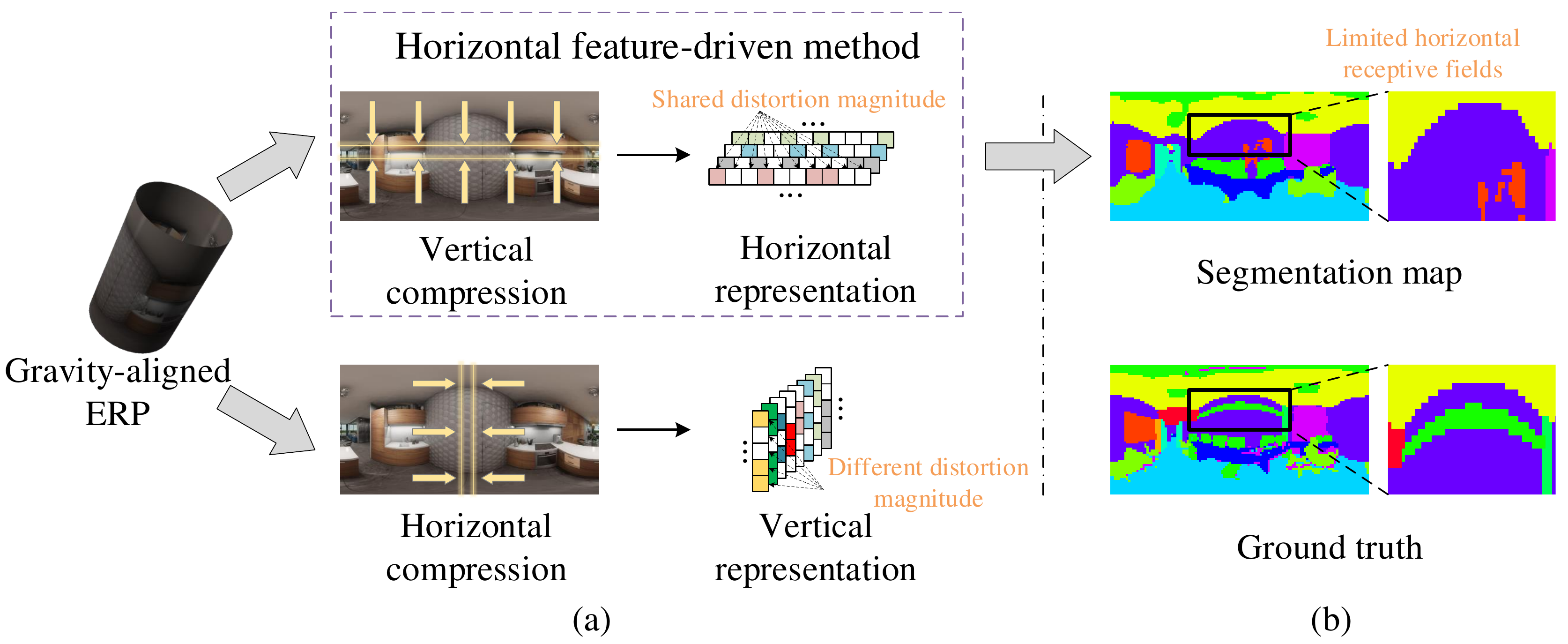}
		\caption{The motivation of the proposed 360° semantic segmentation approach:
			(a) The horizontal representation in each channel shares the same distortion magnitude while the vertical can perceive the distortion distribution.
			(b) The horizontal representation is limited by local receptive fields.}
		\label{motivation}
	\end{figure}

	Most recently, inspired by the geometric nature of gravity-aligned panoramas, some horizontal representation-driven methods \cite{pintore2021slicenet}, \cite{sun2021hohonet} are proposed to address the above problems.
	They squeeze the ERP image into 1D vector along the vertical direction, making it to be more content-focused, as shown in Fig.\ref{motivation}a (top).
	Such a manner can be regarded as a process of shrinking spherical data towards the equator.
	Thus each element in the representation shares the same distortion magnitude and removes the negative effects of panoramic distortions effectively.
	However, due to the fixed compression direction, it is inherently limited to local receptive fields which lack receptive capability in the horizontal direction (Fig.\ref{motivation}b).
	Additionally, if there is no extra guidance during the decoding stage, it will lead to the lack of distortion distribution information in panoramic segmentation results, resulting in unsatisfactory performance.

	Motivated by the horizontal features, we observe that compressing spherical data in the horizontal or vertical direction yields different data representations.
	Essentially, extracting the vertical feature along the horizontal direction is to contract panoramic images towards a certain meridian.
	Considering the data at the same latitude shares the same magnitude of distortion, this operation gathers data belonging to the same magnitude.
	Despite this representation may blur the image content, it makes feature distortions more prominent which can provide implicit distortion distribution prior.
	Compared with existing works that devote to eliminate the effects of distortions, we introduce implicit distortion information to guide the panoramic segmentation.
	Furthermore, the vertical representation also enhances the receptive capability in the horizontal direction.

	In this paper, we present a novel neural network for 360° semantic segmentation consisting of three parts: a feature extraction module, a bi-directional compression module, and an ensemble decoding module.
	To be more specific, we first extract multi-scale features from an ERP image.
	Then our bi-directional compression module encodes the features into two complementary 1D flattened sequences.
	To achieve it, we design a Mix-MLP layer to yield a useful representation before we contract the dimensions.
	Subsequently, we propose a pyramid pooling compression (PPC) layer to perceive both distortion and content information by aggregating different sub-regions with different receptive fields.
	During the ensemble decoding process, we adopt \emph{A-Conv} \cite{pintore2021slicenet} to stretch dimensions and rebuild two different 2D features.
	Finally, the different features are fused for complementarity to predict the segmentation results.

	However, considering the difference between two representations, the feature domains diverge severely, making them difficult to integrate harmoniously.
	Here, we address it by designing a unique self distillation strategy \cite{zhang2019your}.
	Specifically, We divide self distillation into three parallel ones: \emph{horizontal-driven branch (HDB)}, \emph{vertical-driven branch (VDB)}, and \emph{ensembled branch (EB)}, of which \emph{EB} is the fusion result of two representations.
	\emph{HDB} and \emph{VDB} are regarded as student models, while EB is the teacher model. The knowledge in the fusion portion will be shared with the separate portions.
	Finally, the retuned features in student models will feedback to the teacher model, which enhances the interaction of different features and further improves the performance.


	With abundant experiments, we verify that the proposed solution can significantly outperform state-of-the-art algorithms in panoramic semantic segmentation and achieve at least 10\% improvement on the quantitative evaluation.
	Besides, the ablation studies also reveal the effectiveness of our bi-directional representations and specially designed self distillation.
	In summary, our principle contributions are summarized as follows:
	\begin{itemize}
		\item To enlarge the limited horizontal receptive fields and offer implicit distortion prior, we combine horizontal and vertical representations to establish a novel 360° semantic segmentation network from a complementary perspective.
		\item To facilitate the fusion of bi-directional representations, we design a unique self distillation strategy to enhance the interaction of different feature and further improve the performance.
		\item Experiments demonstrate that our method significantly outperforms the current state-of-the-art approaches at least 10\% improvement on all metrics.
	\end{itemize}

	\section{Related Work}

	\subsection{Semantic Segmentation of Omnidirectional Images}
	
	With the progress of 360° camera devices, a wide scene context can be captured in single imagery and thus we can focus on the 360° semantic perception.
	Early approaches \cite{budvytis2019large}, \cite{xu2019semantic} were based on the synthesized panoramic dataset or manually labeled samples to build semantic segmentation systems.
	Motivated by the style transfer \cite{zhu2017unpaired} and data distillation \cite{radosavovic2018data}, Yang \emph{et al.} proposed a framework \cite{yang2019pass}, \cite{yang2020ds}, \cite{yang2021capturing} for re-using the models trained on perspective images by dividing the ERP into multiple restricted FoV sections for predictions.
	Although quite accurate, their strategy relies on labeled perspective image datasets with similar categories and scenes.
	However, recent works find their solutions directly on the real-world datasets \cite{armeni2017joint}.
	Tateno \emph{et al.} \cite{tateno2018distortion} presented spherical convolution filters to make the network aware of the distortion from ERP images.
	Compared to solutions that operate directly on ERP, \cite{zhang2019orientation} projected spherical signals into subdivided icosahedron mesh to mitigate distortion as well as improve prediction accuracy.
	But these methods need to redesign some convolution operations and the performance drops dramatically at the sub-patches boundaries.
	Eder \emph{et al.} \cite{eder2020tangent} introduced tangent images, a novel representation that renders the image onto narrow FoV images tangent to a subdivided icosahedron.
	While solving the above problems, it suffers from complex preprocess and low inference speed.
	Sun \emph{et al.} \cite{sun2021hohonet} used compressing method to encode latent features and use discrete cosine transform (DCT) to finish holistic scene modeling.
	
	\subsection{Horizontal Representation}
	
	Unlike the most existing methods that use pure 2D features to perform prediction, exploiting 1D horizontal representation can make the network learn the underlying geometric correlated knowledge.
	Su \emph{et al.} \cite{su2017learning} utilized different kernel sizes of the standard convolution to overcome the distortions.
	Particularly, the weight can only be shared along the horizontal.
	With the assumption that the horizontal dimension contains rich contextual information, Yang \emph{et al.} \cite{yang2021capturing} proposed a horizontal-driven attention method to capture omni-range priors in 360° images.
	Particularly, Sun \emph{et al.} \cite{sun2021hohonet} used 1D horizontal representation to design HorizonNet for the task of estimating room layout.
	This trend prompts a variety of works on scene understanding.
	For instance, Pintore \emph{et al.} \cite{pintore2021slicenet} proposed SliceNet and adopt Long Short-Term Memory (LSTM) to model the long-range dependencies for 360° depth estimation.
	However, these methods do not consider the latitudinal distortion property and horizontal respective capability, thus leading the accuracy degradation.
	Our solution solves this issue by integrating horizontal and vertical representations simultaneously which we believe is the optimal manner to eliminate the influence of distortions and preserve details.
	Moreover, adding extra tensors will not increase the model complexity and computational cost greatly, because our complexity changes from $ \mathcal{O} \left( W \right)  $ to $ \mathcal{O} \left( W + H \right)  $.
	
	\subsection{Knowledge Distillation}
	
	Knowledge distillation \cite{hinton2015distilling} is one of the most popular compression approaches.
	It is inspired by knowledge transfer from teachers to students.
	And it has shown its superiority in other domains such as data argumentation \cite{bagherinezhad2018label}, adversarial attack \cite{papernot2016distillation}, and model transfer \cite{gupta2016cross}.
	However, it requires substantial efforts and experiments to build teacher models, and we will spend many datasets and long training time to refine student models.
	To overcome the setbacks of traditional distillation, Zhang \emph{et al.}\cite{zhang2019your} proposed a novel training technique named self distillation, which means student and teacher models come from the same networks.
	Therefore, to facilitate the fusion of bi-directional representations, we redesign this technique to adapt to our framework.
	Specifically, we regard two 1D representations as student models, while their fusion results as teacher models, and introduce three loss functions for optimization.
	The well-designed self distillation can enhance the interaction of different feature and further improve the performance.

	\begin{figure}[t]
		\centering
		\includegraphics[width=\linewidth]{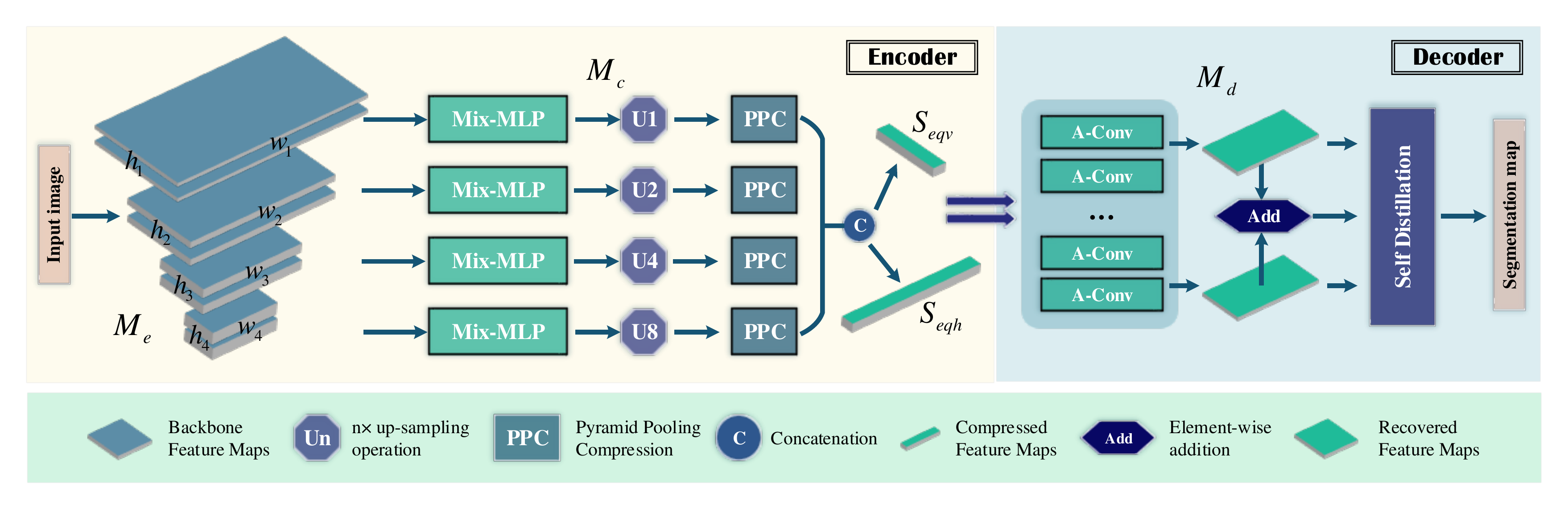}
		\caption{The architecture of the proposed network. 
			This network consists of a feature extraction module $ M_{e} $, a bi-directional compression module $ M_{c} $ and a ensemble decoding module $ M_{d} $.
			Note that the input of four parts in $ M_{c} $ have different resolutions, we draw them at the same size for a convenience exhibition.
		}
		\label{overview}
	\end{figure}

	\section{Approach}

	In this section, we describe the details of the proposed method for 360° semantic segmentation.
	We first show the overview of our framework. 
	Then, the bi-directional compression module that reduces the dimensions of feature maps along horizontal and vertical directions is discussed. 
	Finally, to foster the fusion of bi-directional representations and narrow the feature domain gap, we design a special self distillation strategy to adapt our network structure in the ensemble decoding module.

	\subsection{Network Overview}
	
	The framework is depicted in Fig.~\ref{overview}.
	In our feature extraction module, $ M_{e} $, an ERP format 360° image with the size of $ H \times W  $ will be passed into a deep convolutional neural network, such as \emph{ResNet} \cite{he2016deep}, to progressively decrease the panorama resolution and produce hierarchical feature maps at \{1/4, 1/8, 1/16, 1/32\} of the original image resolution.
	Then we adopt feature pyramid network \cite{lin2017feature} to form multi-scale features, denoted as $ \{F_{i}^{h_{i} \times w_{i}}\}_{i=1,2,3,4}$.
	In the next step, these feature maps are fed into the bi-directional compression module $ M_{c}$ in parallel which contains a lightweight Mix-MLP layer to yield useful representations and a PPC layer to contract the dimensions in the horizontal and vertical directions.
	In particular, we use different pooling operations for different feature maps to aggregate local and global context information.
	We detail this module in Sec.\ref{mc}.
	Then we concatenate multi-level 1D tensors in a single sequence and obtain two representations: $ S_{eqh}$ and $S_{eqv} $. 

	During the decoding period, an ensemble decoding module $ M_{d} $ is employed to reconstruct 2D dense feature maps in Sec.\ref{md}.
	Finally, the fused features are fed into the segmentation head to predict the final segmentation results.
	Besides, for most padding operations, we use circular padding for the left-right boundaries of the feature maps.

	\subsection{Bi-directional Compression Module} \label{mc}
	
	To obtain the bi-directional spherical representations in a more effective and efficient way, we introduce a bi-directional compression module. This module compresses features into two complementary low-dimensional representations which provide content perception and distortion perception separately.
	
	\subsubsection{Mix-MLP Layer.}
	
	Considering the unique structure of ERP, we argue that location information is necessary for our 360° semantic segmentation.
	However, due to the consistent resolution requirements during training and testing, the positional encoding \cite{dosovitskiy2020image} suffers from the inflexible extension problem.
	To enable our network the capability of size-free positional encoding, we present a lightweight Mix-MLP layer.
	Inspired by \cite{chu2021conditional}, \cite{xie2021segformer}, our Mix-MLP mixes a $ 3 \times 3 $ convolution with zero padding and two MLP into a unified framework to introduce implicit location information.
	It can be formulated as:
	\begin{align}
		F_{out} & = Linear \left( \delta \left( DWConv \left( Linear \left( F_{in} \right) \right) \right) \right) + F_{in}
	\end{align}
	where $ F_{in} $ is the feature maps from the backbone, and $ \delta $ is activation function, we use GELU in our experiments.
	The number of channels in $ Linear $ is four times as input.
	Note that we exploit depth-wise convolutions (DWConv) for improving efficiency and reducing the number of parameters.
	As a result, our backbone-extracted features with location information are useful for bi-directional feature compression.

	\begin{figure}[t]
		\centering
		\includegraphics[width=0.98\linewidth]{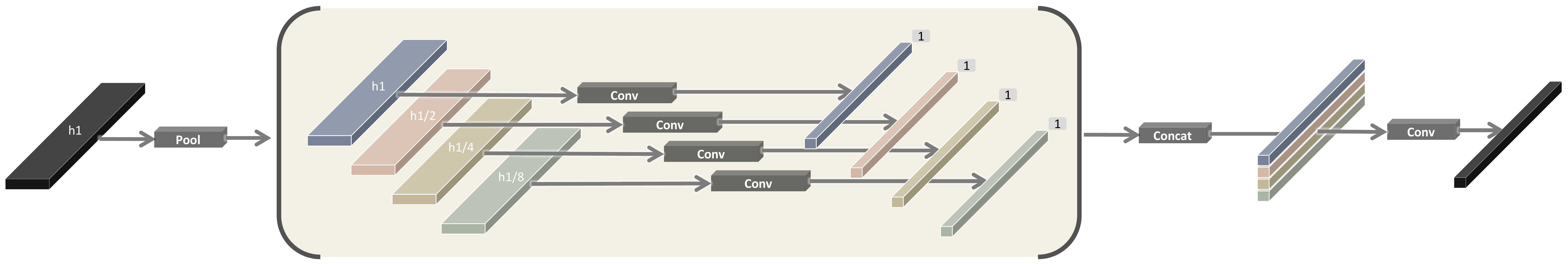}
		\caption{Illustration of pyramid pooling compression (PPC). Given a feature map $ F_{1} $ with the height of $ h_{1} $, to generate horizontal representation, we first use global average pooling to harvest different sub-region representations, then a Conv2D layer is applied to compress the height to 1, followed by concatenation and convolutions layers to form the final 1D representation, which carries both local and global context information. Note that the sizes of sub-regions only vary in the vertical direction.}
		\label{ppcm}
	\end{figure}

	\subsubsection{Pyramid Pooling Compression.}
	
	To squeeze the height $ \left( h \right) $ and width $ \left( w \right) $, the most straightforward operation is to conduct two Conv2D layers with the kernel sizes of $ h \times 1 $ and $ 1 \times w $.
	However, although the receptive field of ResNet is already enough for the works that utilize the 2D feature, it is shown that it is still small for our method that only uses the 1D representations.

	Having observed that the global average pooling is a helpful method as the global contextual prior \cite{liu2015parsenet}, we design an efficient compression method to overcome the above problems.
	Moreover, it is more reasonable to introduce a more powerful representation using sub-regions with different sizes instead of the same size \cite{zhao2017pyramid}.
	Hence, our PPC layer fuses feature under several different pyramid scales, Fig.\ref{ppcm} gives an example.
	Note that the number of pyramid levels and size of each level can be modified manually. 
	They are related to the size of the feature maps that are fed into the pooling layer.
	Therefore, combined with our hierarchical structure, the number of pyramid levels decreases with the increase of the network stage.
	Concretely, given the feature maps $ \{F_{i}^{h_{i} \times w_{i}}\}_{i=1,2,3,4} $, the pooling size is $ \{ \frac{h_{i}}{2^{j}} \times w_{i} \}_{i=1,2,3,4; j=0,...,4-i} $ for horizontal features, and $ \{ h_{i} \times \frac{w_{i}}{2^{j}} \}_{i=1,2,3,4; j=0,...,5-i} $ for vertical features, respectively.

	Besides, because the feature maps in different stages have different sizes, extra upsampling layers (see Fig.\ref{overview}) should be added to align them.
	For example, we only upsample the feature maps along the horizontal direction to align the horizontal feature. 
	By compressing the features in different directions, our model can implicitly perceive content information and distortion distribution in a panorama from two perspectives.

	\subsection{Ensemble Decoding Module} \label{md}
	
	To produce per-pixel predictions from 1D representations, Sun \emph{et al.} \cite{sun2021hohonet} exploited interpolation operations and inverse discrete cosine transform (IDCT), leading to reduction of the model learnability.
	Different from the strategy of reshaping the size directly, we utilize $ n $ \emph{A-Conv} layers \cite{pintore2021slicenet}, each of which includes an upsampling layer, a Conv2D layer, a BN layer, and a PReLU, to progressively stretch dimensions.
	Note that we replace the PReLU with ReLU for our segmentation modalities, and the final resolution is $ \frac{H}{4} \times \frac{W}{4} $. In this way, we can obtain two distinct reconstructed tensors, denoted as $ D_{h} $ and $ D_{v} $.

	\begin{figure}[t]
		\centering
		\includegraphics[width=0.98\linewidth]{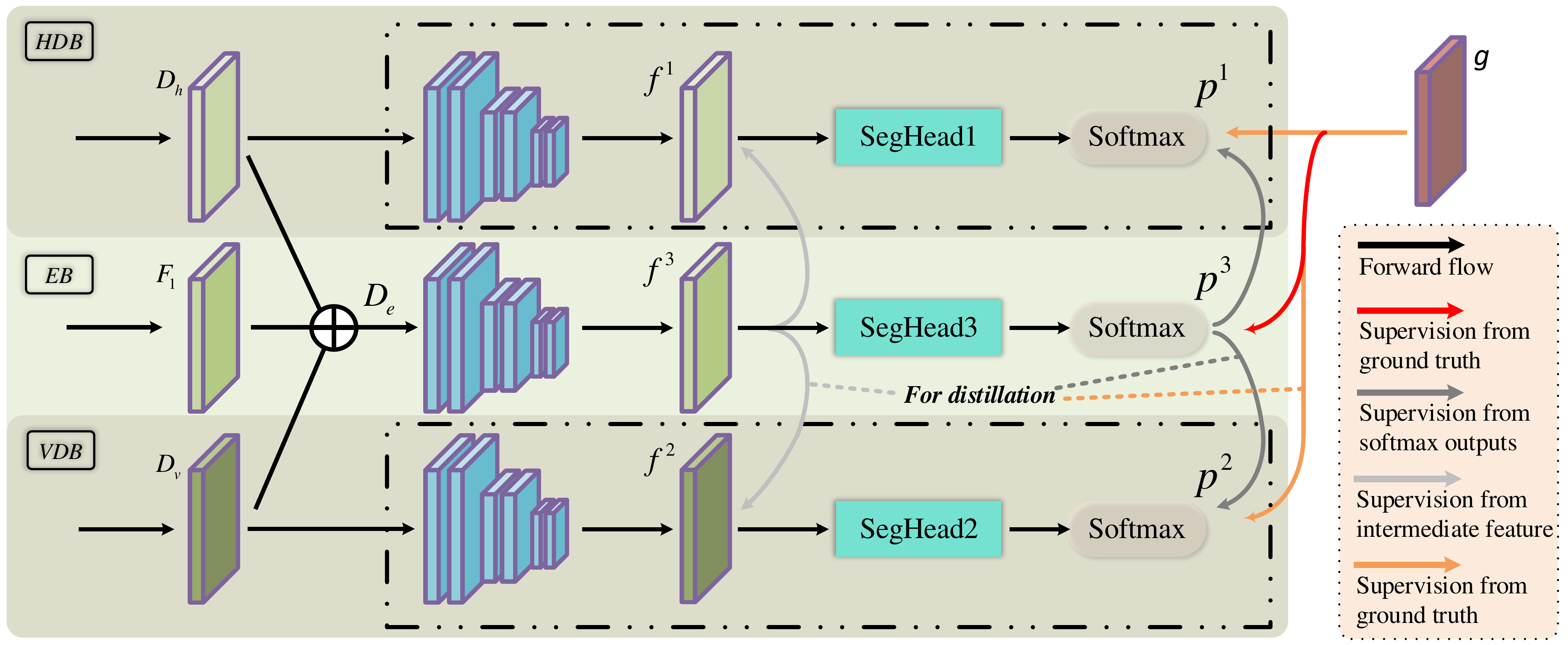}
		\caption{
			Illustration of our self distillation strategy.
			During the network's training process, we principally divide our self distillation structure into three sections according to their sources.
			The $ D_{e} $ is the summation of $ D_{h} $, $ D_{v} $, and $ F_{1} $.
			In the test stage, these structures in rectangle boxes which only be introduced in training processes will be removed.}
		\label{self distillation}
	\end{figure}

	\subsubsection{Self Distillation.}
	
	Despite we can directly ensemble these features to capture complementary semantic information in horizontal and vertical directions, the direction-dominated characteristics are still underutilized.
	Furthermore, due to bi-directional features representing different panoramic characteristics, the feature domain gap will hinder the fusion process.
	
	To facilitate the fusion of different features, we do not design a more complex fusion network that can significantly enlarge the model size. On the contrary, we design a unique self distillation strategy in the decoding stage to narrow the domain gap between different features. As illustrated in Fig.~\ref{self distillation}, our self distillation structure can be divided into three parts: \emph{HDB}, \emph{VDB}, and \emph{EB}. 
	The \emph{EB} comprises \emph{HDB}, \emph{VDB}, and the backbone feature $ F_{1}^{h_{1} \times w_{1}} $.
	During the training process, \emph{HDB} and \emph{VDB} are regarded as student models while \emph{EB} is the teacher model. The students can learn beneficial knowledge from the teacher, and the teacher can obtain good feedback from the students. In this manner, both students and teacher can benefit from each other. 
	We set several convolution layers (bottleneck) and a SegHead as the segmentation part, and both students and teacher share the same network structure.
	The bottleneck contains three Conv2D layers with kernel sizes of $ 1 \times 1  $, $ 3 \times 3  $, and $ 1 \times 1  $.
	The SegHead comprises two upsampling layers to reach the original resolution, and two Conv2D layers to predict the segmentation mask at a $ H \times W  \times N_{cls} $ resolution, where $ N_{cls} $ is the number of categories. 
	Due to page limitations, we exhibit a complete example in the supplementary material.
	
	After predicting the segmentation map of a 360° image, it is straightforward to adopt the segmentation labels to supervise the \emph{HDB} and \emph{VDB}, which can produce better $ D_{h} $ and $ D_{v} $.
	Nevertheless, if we only use this supervision, the knowledge will not interact between the students and teacher.
	To this end, we introduce two extra supervisions (supervisions from intermediate features and final softmax outputs of the teacher model) to encourage the student models to learn from the teacher model. 
	In brief, we use cross-entropy loss, kullback-Leibler(KL) divergence loss, and L2 loss as the optimization functions in our self distillation strategy.

	\subsection{Objective Function}

	
	
	
	Our objective function comprises three kinds of loss as the objective functions for optimizing predictions. 
	
	\paragraph{Cross-Entropy Loss.} The first supervision is the cross-entropy loss.
	Almost all CNNs for this task exploit cross-entropy loss. 
	It is computed with the ground truth (GT) from the training samples and the predictions of the softmax layer. 
	We deploy it not only to the teacher's branch but also to two student branches.
	Through cross-entropy loss, the knowledge hidden in the training set is introduced directly from GT to all the branches.
	It can be written as:
	
	\begin{align}
		L_{ce} &= \langle -g\log(p^{i}) \rangle
	\end{align}
	where  $ g $, $ p^{i} $ denote the GT and predicted values from softmax layer’s output, respectively;
	$ i \in \{ 1, ..., N \} $, where $ N $ denotes the number of SegHeads in the training period, and $ N=3 $ in our experiments (refer to Fig\ref{self distillation}).
	Moreover, class-wise weighted \cite{zhang2019orientation} is utilized to balance different classes examples.
	
	\paragraph{KL Divergence Loss.} The second supervision is the KL divergence loss.
	We use KL divergence to measure the difference between two distributions.
	It can be obtained through the computation of softmax outputs between students and teachers.
	Under the teacher’s guidance, the distributions of students' SegHeads can approximate the teacher's, which indicates the supervision from distillation.
	It can be obtained by:
	
	\begin{align}
		L_{kl} &= \langle p^{N}\log(\dfrac{p^{N}}{p^{i}}) \rangle
	\end{align}
	
	\paragraph{L2 Loss.} The last supervision is $ L_{2} $ loss which works by decreasing the distance between feature maps in the student branches and the teacher branch. 
	In this way, the knowledge in feature maps is distilled to each student's bottleneck layer.
	
	\begin{align}
		L_{l2} &= \Arrowvert f^{i} - f^{N} \Arrowvert _{2}^{2}
	\end{align}
	
	Note that the last two losses for the teacher model are zero, which means the supervision in the teacher model only comes from GT. 
	Most importantly, we denote it as base loss $L_{b}$  in our network without distillation.
	For student models, we collect all supervision to obtain self distillation loss $L_{s}$.
	Meanwhile, to make the fusion process more interactive, we adopt three hyper-parameters to balance them.
	
	\begin{align}
		\begin{split}
			L_{total} &= L_{b} + L_{s}, \\
			&= \sum_{i=N} L_{ce} + \sum_{i=1}^{N-1} (\alpha * L_{ce} + \beta * L_{kl} + \gamma * L_{l2}).
		\end{split}
	\end{align}

	\section{Experiments}

	We evaluate the effectiveness of our model in this section by carrying out comprehensive experiments on a real-world dataset.
	In the following subsections, we first introduce the dataset and implementation details, then report quantitative and qualitative results compared with the state-of-the-art approaches.
	Finally, we perform a series of ablation experiments for the proposed components.

	\subsection{Dataset and Evaluation Metrics}

	We evaluate our method on the Stanford 2D-3D-S dataset \cite{armeni2017joint}, which consists of 1413 real-world equirectangular RGB-D images over 13 categories. 
	The dataset contains six large-scale indoor areas and provides semantic labels with the ERP format as GT annotations.
	Furthermore, the panoramas have a resolution of 2048 $\times$ 4096 and contain black void regions at the top and bottom.
	Following the prior works, we report averaged quantitative results from the 3-fold cross-validation splits.
	
	We adopt Mean IoU (mIoU, mean of class-wise intersection over union) and Mean Acc (mAcc, mean of class-wise accuracy) as evaluation metrics for the task of 360° semantic segmentation.

	\subsection{Implementation Details}

	We conduct our experiments on three resolutions: 64 × 128, 256 × 512, and 512 × 1024.
	We train our learning model using Adam \cite{kingma2014adam} optimizer on a GTX 3090 GPU, and the batch sizes are set to 16, 8, and 4.
	For the low-resolution (the first two) inputs, we use the residual UNet-style architecture as backbone \cite{cohen2019gauge}, \cite{jiang2018spherical}, \cite{zhang2019orientation} and replace the specialized kernels with planar one.
	For the high-resolution (the last one) inputs, we adopt ResNet-101 pre-trained on COCO \cite{lin2014microsoft} as backbone \cite{eder2020tangent}, \cite{sun2021hohonet} to capture the larger receptive field.
	Inspired by \cite{chen2017deeplab}, \cite{huang2019ccnet}, we employ the poly learning rate policy where the base learning rate is multiplied by $ (1-\frac{iter}{max\_iter})^{power} $ with $ power=0.9 $.
	The learning rate is set to $ 1 \times 10^{-3} $ with $ max\_iter=300 $ for low-resolution and $ 1 \times 10^{-4} $ with $ max\_iter=60 $ for high resolution.
	To prevent overfitting, we adopt a simple augmentation strategy of randomly cutting a patch of the input image and padding this region with a black mask, where the sizes of the hole are chosen from the set \{20 $\times$ 40, 80 $\times$ 160, 160 $\times$ 320\}. In our objective function, we set $ \alpha=0.7 $, $ \beta=0.3 $, and $ \gamma=0.003 $.

	\setlength{\tabcolsep}{4pt}
	\begin{table}[t]
		\begin{center}
			\renewcommand{\arraystretch}{1.05}
			\caption{Quantitative evaluation on Stanford2D3D dataset.
				Note that the results are averaged over the 3-folds.
				Reasons for different high resolutions, refer to Sec.\ref{results}.}
			\label{quantitative results}
			\begin{tabular}{cccc||cc}
				\hline
				$ H \times W $ & Input & Method & Pub. \& Year & mIoU $ \uparrow $ & mAcc $ \uparrow $ \\
				\hline\hline
				\multicolumn{6}{l}{Low-resolution input} \\
				\hline
				\multirow{7}{*}{64 $\times$ 128} 
				& RGB-D & Gauge Net \cite{cohen2019gauge} & ICML'19 & 39.4 & 55.9 \\
				& RGB-D & UGSCNN \cite{jiang2018spherical} & ICLR'19 & 38.3 & 54.7 \\
				& RGB-D & HexRUNet \cite{zhang2019orientation} & ICCV'19 & 43.3 & 58.6 \\
				& RGB-D & SWSCNN \cite{esteves2020spin} & NeruIPS'20 & 43.4 & 58.7 \\
				& RGB-D & TangentImg \cite{eder2020tangent} & CVPR'20 & 37.5 & 50.2 \\
				& RGB-D & HoHoNet \cite{sun2021hohonet} & CVPR'21 & 40.8 & 52.1 \\
				& RGB-D & Ours & - & \textbf{47.2} & \textbf{61.2} \\
				\hline
				\multirow{3}{*}{256 $\times$ 512} 
				& RGB-D & TangentImg \cite{eder2020tangent} & CVPR'20 & 41.8 & 54.9 \\
				& RGB-D & HoHoNet \cite{sun2021hohonet}  & CVPR'21 & 43.3 & 53.9 \\
				& RGB-D & Ours & - & \textbf{53.8} & \textbf{66.5} \\
				\hline\hline
				\multicolumn{6}{l}{High-resolution input} \\
				\hline
				$ 2048 \times 4096 $ & RGB & TangentImg \cite{eder2020tangent} & CVPR'20 & 45.6 & 65.2 \\
				$ 1024 \times 2048 $ & RGB & HoHoNet \cite{sun2021hohonet} & CVPR'21 & 52.0 & 65.0 \\
				$ 512 \times 1024 $ & RGB & Ours & - & \textbf{52.2} & \textbf{65.6} \\
				\hline
				$ 2048 \times 4096 $ & RGB-D & TangentImg \cite{eder2020tangent} & CVPR'20 & 51.9 & 69.1 \\
				$ 1024 \times 2048 $ & RGB-D & HoHoNet \cite{sun2021hohonet} & CVPR'21 & 56.3 & 68.9 \\
				$ 512 \times 1024 $ & RGB-D & Ours & - & \textbf{56.7} & \textbf{70.8} \\
				\hline
			\end{tabular}
		\end{center}
	\end{table}

	\subsection{Results and Analysis} \label{results}

	\subsubsection{Compared with State-of-the-arts}
	
	In this subsection, we compare our method with the state-of-the-art methods on 360° semantic segmentation in both quantitative and qualitative evaluations, for which the numerical results or segmentation map on the same dataset is available.
	
	\paragraph{Quantitative Evaluation:} Table.\ref{quantitative results} shows the quantitative comparison results with the current state-of-the-art methods on different input resolutions.
	It is evident that our approach substantially outperforms the compared approaches in all metrics.
	From these evaluations on the lowest resolution, we can conclude that:
	
	(i) Compared with the spherical CNNs methods \cite{cohen2019gauge}, \cite{esteves2020spin} \cite{jiang2018spherical} which aim to directly learn distortion-aware representation from the sphere, our approach avoids complex convolutional design on the transfer between planar and sphere, showing more promising generality and flexibility.
	
	(ii) Compared with the distortion-tolerate approaches \cite{eder2020tangent}, \cite{zhang2019orientation}, which project 360° images into icosahedron format, our approach only need ERP as input and omit the process of transformation.
	For example, our approach outperforms HexRUNet \cite{zhang2019orientation} which equip a specially non-rectangular kernel by a significant margin, with approximately 9\% improvement on mIoU and 4\% improvement on mAcc.
	
	(iii) As benefits of introducing vertical representation which provides guidance of distortion distribution and enlarges receptive fields in horizontal direction during the learning stage, our approach achieves a 17\% improvement on both mIoU and mAcc on the lowest resolution compared with the Sun \emph{et al.} \cite{sun2021hohonet} that only use horizontal representation.
	
	To further demonstrate the generality of our method, we conduct experiments on other resolutions.
	It can be observed that our network has achieved satisfactory results on the 256 $ \times $ 512 resolution with at least 24\% improvements on mIoU and 23\% on mAcc.
	Unfortunately, due to the limitation of our device, we failed to train our network on higher resolutions as \cite{eder2020tangent} and \cite{sun2021hohonet}.
	However, based on the fact that the segmentation performance is positively correlated with the resolution size, we train our network with ResNet-101 as the backbone on a lower 512 × 1024 resolution and still achieve competitive performance.

	\begin{figure}[t]
		\centering
		\includegraphics[width=\linewidth]{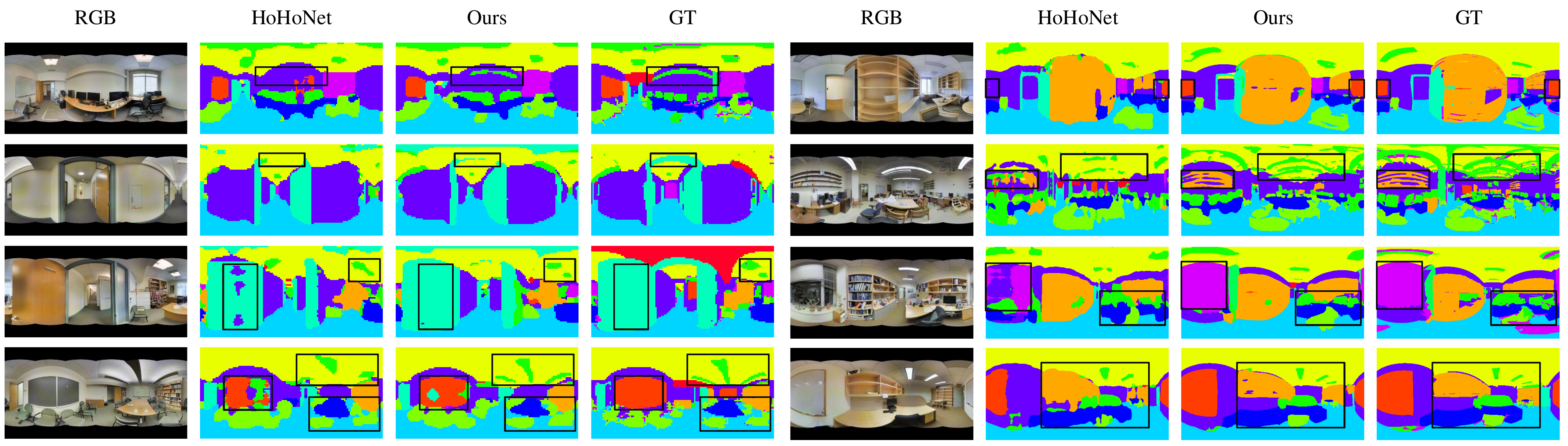}
		\caption{Qualitative evaluations of the segmentation results on 64 $ \times $ 128 (left) and 256 $ \times $ 512 (right) resolutions.
			Black rectangles are used to highlight difference.}
		\label{segmentation map}
	\end{figure}

	\paragraph{Qualitative Evaluation:} Fig.\ref{segmentation map} shows qualitative results on Stanford2D3DS dataset, compared to HoHoNet \cite{sun2021hohonet}.
	From the figure, we can observe that our approach performs well on all indoor scenes, while the horizontal representation method shows inferior segmentation results, especially in regions with distortions or regions with complex contextual information.
	With the complementary relationship between two representations, our method has a larger receptive field and sufficient distortion information.
	For example, the class with a strong distribution along horizontal direction while weaker along the vertical direction (see Fig.\ref{segmentation map} (left) first two rows) has an inferior performance in HoHoNet.
	Because these pixels occupy a small proportion in each column, they will be omitted when compressing the height dimension and are difficult to recover.
	In contrast, our vertical representation perceives this distribution in another dimension and supplement it to the decoding module.
	In general, our approach achieves a better performance from local details (receptive fields) to global distribution (distortion shape), which benefits from the designed modules. More qualitative results are included in the supplementary material.

	\setlength{\tabcolsep}{3.65mm}
	\begin{table}[t]
		\begin{center}
			\renewcommand{\arraystretch}{1.0}
			\caption{Ablation study with the key components on our 360° semantic segmentation approach.
				Experiment resolution: 64 $ \times $ 128.}
			\label{ab: hvmp}
			\begin{tabular}{cc|cc||cc}
				\hline
				$ \mathcal{H} $ & $ \mathcal{V} $ &  $ \mathcal{M} $ & $ \mathcal{P} $ & mIoU & mAcc \\
				\hline\hline
				\Checkmark & \XSolidBrush & \XSolidBrush & \XSolidBrush & 41.50 & 53.27 \\
				\XSolidBrush & \Checkmark & \XSolidBrush & \XSolidBrush & 40.63 & 52.65 \\
				\Checkmark & \Checkmark & \XSolidBrush & \XSolidBrush & 42.76 & 55.00 \\
				\Checkmark & \Checkmark & \Checkmark & \XSolidBrush & 43.35 & 55.84 \\
				\Checkmark & \Checkmark & \Checkmark & \Checkmark & \textbf{44.71} & \textbf{57.03} \\
				\hline
			\end{tabular}
		\end{center}
	\end{table}

	\subsubsection{Ablation Studies}
	
	To validate the effectiveness of different components in our approach, we conduct ablation studies and as illustrated in Table.\ref{ab: hvmp} and Table.\ref{ab: sd}.
	Note that all the experiment results are evaluated on the lowest resolution input.
	
	\paragraph{Effectiveness of bi-directional representations:} We remove the self distillation to explore the effectiveness of combining two representations (horizontal ($ \mathcal{H} $), vertical ($ \mathcal{V} $)).
	Concretely, we implement our model without other key schemes (Mix-MLP ($ \mathcal{M} $), PPC ($ \mathcal{P} $)) and provide the quantitative results of variants equipped with different representations.
	From Table.\ref{ab: hvmp} (first three rows), we can observe that the joint representation performs better than only using one directional representation, which indicates that our network gains information from two complementary perspectives to facilitate the accuracy.
	In addition, we can observe that the $ \mathcal{H} $ performs well than the $ \mathcal{V} $, which proves that the vertical representation contains implicit distortion prior and blurs the content.
	
	\paragraph{Effectiveness of components in $ M_{c} $:} Subsequently, we gradually add the removed components to show the different segmentation performances.
	Note that we utilize a Conv2D layer to compress dimensions without $ \mathcal{M} $.
	As seen from Table.\ref{ab: hvmp} (last three rows), the mIoU is improved from 42.76 to 44.71 with a percentage gain of 4.6\%, and the mAcc is boosted from 55.00 to 57.03 with the percentage gain of 3.7\%.
	It also can be found that our network with useful position information derived from $ \mathcal{M} $ achieves pleasing results.
	For the compression strategy, $ \mathcal{P} $ provides large receptive fields and sufficient contextual information, making our model gains further improvements and outperforms a single Conv2D layer with 3.1\% on mIoU and 2.1\% on mAcc.
	Finally, the completed framework achieves the best results proving the effectiveness of our proposed components.

	\setlength{\tabcolsep}{0.65mm}
	\begin{table}[t]
		\begin{center}
			\renewcommand{\arraystretch}{1}
			\caption{Ablation study with the self distillation strategy.
				Experiment resolution: 64 $ \times $ 128.}
			\label{ab: sd}
			\begin{tabular}{c|cc|cc|cc||cc|cc|cc}
				\hline
				\multirow{3}{*}{Fold} & \multicolumn{6}{c||}{\emph{w/o} self distillation} & \multicolumn{6}{c}{\emph{w/} self distillation} \\  \cline{2-13}
				& \multicolumn{2}{c|}{\emph{VDB}} & \multicolumn{2}{c|}{\emph{HDB}} & \multicolumn{2}{c||}{\emph{EB}} & \multicolumn{2}{c|}{\emph{VDB}} & \multicolumn{2}{c|}{\emph{HDB}} & \multicolumn{2}{c}{\emph{EB}} \\ \cline{2-13}
				& mIoU & mAcc & mIoU & mAcc & mIoU & mAcc & mIoU &mAcc & mIoU &mAcc & mIoU & mAcc \\
				\hline\hline
				1 & 38.68 & 50.16 & 44.91 & 56.11 & 47.08 & 58.05 & 45.12 & 57.29 & 49.34 & 61.06 & 50.48 & 61.93 \\
				2 & 33.48 & 48.72 & 36.19 & 51.68 & 38.23 & 52.37 & 37.22 & 53.90 & 40.41 & 57.00 & 40.87 & 57.83 \\
				3 & 41.19 & 53.71 & 46.59 & 59.39 & 48.82 & 60.68 & 46.30 & 61.12 & 49.20 & 63.57 & 50.35 & 63.84 \\
				\hline
			\end{tabular}
		\end{center}
	\end{table}

\begin{figure}[t]
		\centering
		\includegraphics[width=\linewidth]{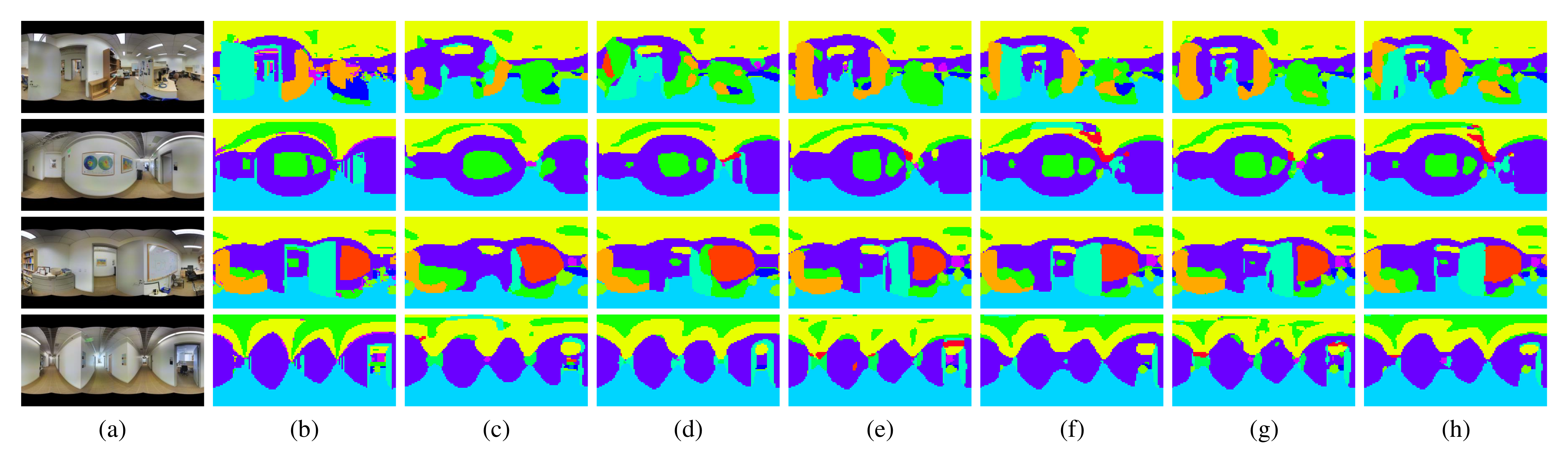}
		\caption{Visual ablation comparison on our 360° semantic segmentation approach. 
			(a) panoramic image.
			(b) ground truth.
			(c) \emph{VDB} \emph{w/o} self distillation.
			(d) \emph{VDB} \emph{w/} self distillation.
			(e) \emph{HDB} \emph{w/o} self distillation.
			(f) \emph{HDB} \emph{w/} self distillation.
			(g) \emph{EB} \emph{w/o} self distillation.
			(h) \emph{EB} \emph{w/} self distillation.
		}
		\label{ab: segmentation map}
	\end{figure}

	\paragraph{Effectiveness of self distillation:} Since the different representations have a severe feature domain gap, it is difficult to integrate them harmoniously.
	Thus the self distillation plays the role of facilitating the fusion of bi-directional representations in our method.
	Furthermore, different from other knowledge distillation methods that pre-training a large teacher model, we exploit self distillation by directly dividing our network into student models (\emph{VDB} and \emph{HDB}) and teacher model (\emph{EB}).
	To validate the effects of this strategy, we experiment with removing all supervision for student models, which means the knowledge from the teacher and dataset are obstructed.
	The quantitative results are shown in Table.\ref{ab: sd}, we report detailed semantic segmentation results on three folds.
	From Table.\ref{ab: sd}, we can conclude that via self distillation, all branches gain a significant improvement, which indicates that the well-designed training technique can foster the interaction of bi-directional representations and notably improve the segmentation performance.
	We also present the qualitative comparison results in Fig.\ref{ab: segmentation map}.
	

	\section{Conclusions}

	In this paper, a novel panoramic semantic segmentation network is presented from a complementary perspective by combining horizontal and vertical representations, which is capable of expanding the limited horizontal receptive fields and offering implicit distortion prior.
	To integrate complementary bi-directional representations, we design a unique self distillation strategy to enhance the interaction of different representations and make the predicted segmentation map more accurate.
	To our knowledge, this is the first work in panoramic visual perception that uses bi-directional feature compression to achieve the complementary.
	As the benefit of the proposed complementary representation, our approach significantly outperforms current state-of-the-art solutions in prediction accuracy on the real-world indoor dataset.
	

	\clearpage
	%
	%
	\bibliographystyle{splncs04}
	\bibliography{egbib}
\end{document}